\documentclass[letterpaper,twocolumn,10pt]{article}
\usepackage{zhanggroup}

\usepackage{hyperref}
\usepackage{url}
\usepackage{booktabs}
\usepackage{amsfonts}
\usepackage{nicefrac}
\usepackage{microtype}
\usepackage{xcolor}
\usepackage{caption,subcaption}
\usepackage{graphicx}
\usepackage{xspace}
\usepackage{multirow}
\usepackage[normalem]{ulem}
\usepackage{amsmath}
\usepackage{wrapfig}
\usepackage{tcolorbox}
\usepackage{etoolbox}
\usepackage{enumitem}
\usepackage{array}
\usepackage{pifont}

\newcommand{\mypara}[1]{\noindent{\bf {#1}.} \xspace}
\newcommand{\meanline}{%
  \raisebox{0.5ex}{%
    \rule{3pt}{0.45pt}\hspace{1pt}%
    \rule{3pt}{0.45pt}\hspace{1pt}%
    \rule{3pt}{0.45pt}%
  }%
}

\newcommand{\medianpoint}{%
  \raisebox{-0.1ex}{$\circ$}%
}
\newcommand{\cmark}{\ding{51}}%

\begin{document}

\date{}

\title{\bf Can Frontier Auto Researchers \\Reconstruct Human Research Decisions in AI?}

\author{
Shengyun Si\ \ \
Junjie Chu\ \ \
Ye Leng\ \ \
Yang Zhang\thanks{Corresponding author}\ \ \
\\
\\
\textit{CISPA Helmholtz Center for Information Security}
}

\maketitle

\begin{abstract}
Scientific research has long been driven by substantial human effort, from reviewing prior work and identifying research gaps to designing specific solutions.
Recent advances in harness-based agent systems have given rise to increasingly capable \emph{Auto Researchers}, which can autonomously carry out these tasks and, increasingly, entire end-to-end research workflows.
Yet despite their rapidly growing capabilities, how closely the research decisions and outputs produced by auto researchers align with those of human experts remains underexplored.
We investigate this gap through the lens of research reconstruction and study: \emph{Given comparable research context, how similar are the research decisions and outputs produced by Auto Researchers to those of human researchers in AI research?}
To answer this question, we introduce \textbf{Auto}Researcher--\textbf{H}uman \textbf{R}esearch \textbf{S}imilarity (\textbf{AutoHRS}), a stage-wise evaluation framework for assessing how faithfully auto researchers reconstruct human research artifacts across different stages of the scientific workflow.
Using 152 ICLR oral papers as our corpus, we generate 456 research trajectories spanning different stages of the research process.
Applying AutoHRS to these trajectories, we find that current \textit{Auto Researchers} frequently fail to recover key research decisions and their resulting artifacts across multiple stages of the research process.
\end{abstract}

\section{Introduction}

\begin{figure*}[!t]
    \centering
    \includegraphics[width=0.95\textwidth]{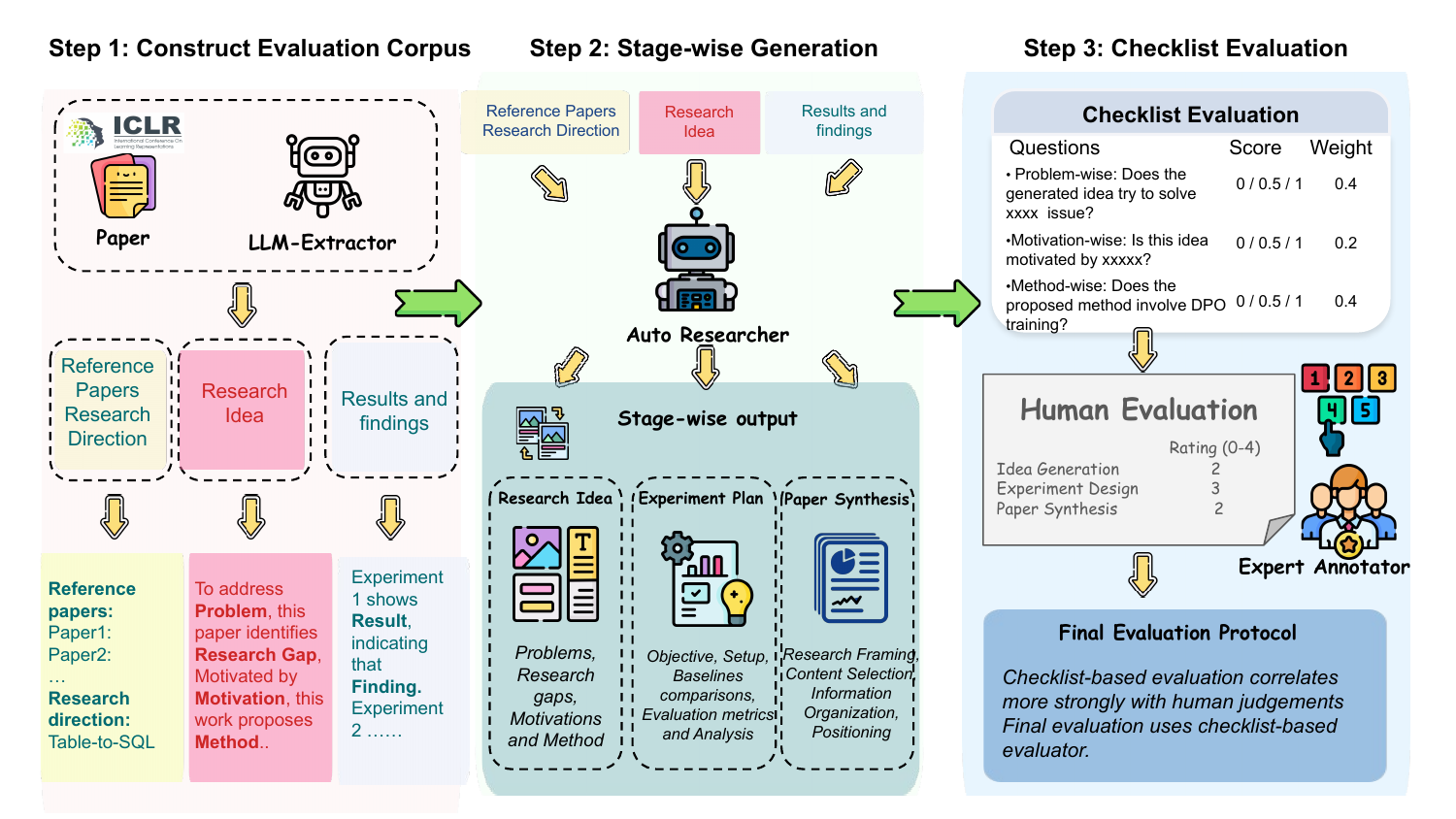}

    \caption{Overview of our work.
    We construct AutoHRS by decomposing human research into three stage-wise tasks: \textit{idea generation}, \textit{experiment design}, and \textit{paper synthesis}. Generated outputs are evaluated against human research using a checklist-based evaluator validated by expert judgments.}

    \label{figure:overview}
\end{figure*}

\raggedbottom
Large language models (LLMs) have been increasingly integrated into scientific research since their emergence, supporting tasks ranging from scientific discovery to scientific writing~\cite{WDJH24, LLLFCH24, BJCH25, GSXHKWBZ25,  KGGE25, LXGZLYZJXDRZFB25, SYH25, SH26, KDLZPXHQSLCZC26}. 
Despite these advances, human researchers remained in control of the overall research process, while LLMs primarily served as assistants for individual tasks. With the development of harness-based agents, however, language models are increasingly being integrated into end-to-end scientific workflows~\cite{YLL26, WXHGWX26, THLCS26}. 
We refer to such harness-based systems that autonomously carry out multiple stages of the research process as \emph{Auto Researchers}. In these emerging systems, the traditional division of labor is substantially altered: auto researchers take a leading role in carrying out the research process, while human researchers increasingly provide guidance, feedback, and oversight~\cite{SSWSWYLMLB25, YLLJ26, ZW26}. 
This shift raises an important question of how the research produced by these systems should be evaluated. 
However, evaluating research independently generated by auto researchers is challenging: its novelty, scientific value, and potential for genuine breakthroughs are difficult to establish without extensive validation. 
A more tractable starting point is to use established human research as a reference for assessing how these systems formulate ideas, design experiments, and synthesize findings into papers. 
We therefore ask: \emph{Can Frontier Auto Researchers Reconstruct Human Research Decisions in AI?}

Existing evaluation frameworks for LLM-based scientific research systems have largely focused on model-level capabilities, typically assessing performance on individual research tasks rather than the behavior of harness-based systems operating across the full research process~\cite{CCNZWYLLWLDXBBAHNGSS25, WLBJPBCCCDEGGJKLNSSTWB25,ZWGC26,XLYCCGWLLXCYZZMGWZYCHLMXZCZGZHXGWLSLZTLSYSWYYFZOZBZ26,FLYCLTCSMD26,CYZXWL26}. 
In this work, rather than evaluating auto researchers solely by whether they can complete predefined research tasks, we ask whether they can conduct research in ways that reconstruct the decisions and processes of human researchers. 
To operationalize this evaluation, we introduce \textbf{Auto}Researcher–\textbf{H}uman \textbf{R}esearch \textbf{S}imilarity (\textbf{AutoHRS}), a stage-wise evaluation framework for assessing how faithfully auto researchers reconstruct human research artifacts\footnote{Reconstructing research artifacts does not mean reproducing the original study; we use reconstruction as a proxy to observe whether auto researchers can mimic human research decisions and outputs.} across different stages of the scientific process. 
As scientific research is an open-ended, nonlinear, and multi-stage process, we decompose it into three tractable stages that can be systematically evaluated: idea generation grounded in reference papers and a research direction, experiment design conditioned on a given research idea, and paper synthesis grounded in provided research context and experimental findings~\cite{LSBZ87, KD88}.

To instantiate this evaluation framework, we construct an evaluation corpus from ICLR 2026 oral papers. Each paper is decomposed into three research trajectories corresponding to three stages considered in AutoHRS: from reference papers to a research idea, from a research idea to an experimental plan, and from experimental findings to the final paper. 
These trajectories provide both the inputs to the auto researchers and corresponding human research outputs used as references for evaluation. 
Our framework employs a fine-grained checklist-based evaluation pipeline, with stage-specific dimensions and weights tailored to stages of the research process.
We further validate the framework against human judgments, showing strong alignment with human judgments.

Our evaluation shows that current auto researchers exhibit limited ability to reconstruct human research across all three stages. 
Under AutoHRS, on a normalized scale from 0 to 1, auto researchers achieve a score of 0.29 for the best-scoring idea per paper in idea generation, 0.36 on experiment design, and 0.65 on paper synthesis. 
To better understand these results, we analyze where and how auto researchers diverge from human research, identifying recurring patterns in idea composition and experiment planning. 
We further analyze factors that affect reconstruction performance, including the granularity of the input research direction and the research type.
These results suggest that, despite their growing ability to automate end-to-end scientific workflows, current auto researchers still differ substantially from human researchers in how they formulate ideas, design experiments, and synthesize research into scientific papers.

\section{AutoHRS Evaluation Framework}

We propose an evaluation framework for auto researchers that assesses whether they can reconstruct the research decisions made by human researchers. 
In this framework, we define three stage-wise evaluation tasks corresponding to three key stages of the research process: literature-grounded idea generation, idea-grounded experiment design, and findings-grounded paper synthesis. 

\subsection{Stage-Wise Evaluation Tasks}
\label{section:stage-wise}

\mypara{Literature-Grounded Idea Generation} 
In this task, each instance consists of a set of reference papers, $\mathcal{R}=\{R_1,R_2,\ldots,R_n\}$, and a general research direction $D$. 
Each reference paper $R_i$ is represented by five components, $R_i=\{t_i,a_i,p_i,m_i,o_i\}$, corresponding to its \textit{title}, \textit{abstract}, \textit{problem}, \textit{method}, and \textit{results}, respectively. 
Given the reference set $\mathcal{R}$ and research direction $D$, the target output is a structured research idea, represented as $I=\{p,g,r,m\}$, where $p$ denotes the research problem, $g$ the research gap, $r$ the research motivation, and $m$ the proposed method.

\mypara{Idea-Grounded Experiment Design}
In this stage, the auto researcher is given the structured research idea $I=\{p,g,r,m\}$ defined above. 
Given the input $I$, the auto researcher is tasked with producing a detailed experiment plan, denoted as $P$, that designs experiments around the proposed method to evaluate whether it addresses the identified research problem and gap. 
The plan is expected to specify the main experimental objectives, evaluation settings, datasets or benchmarks, comparison methods, evaluation criteria, and additional analyses needed to test the proposed idea.

\begin{figure*}[!t]
    \centering
    \includegraphics[width=0.95\textwidth]{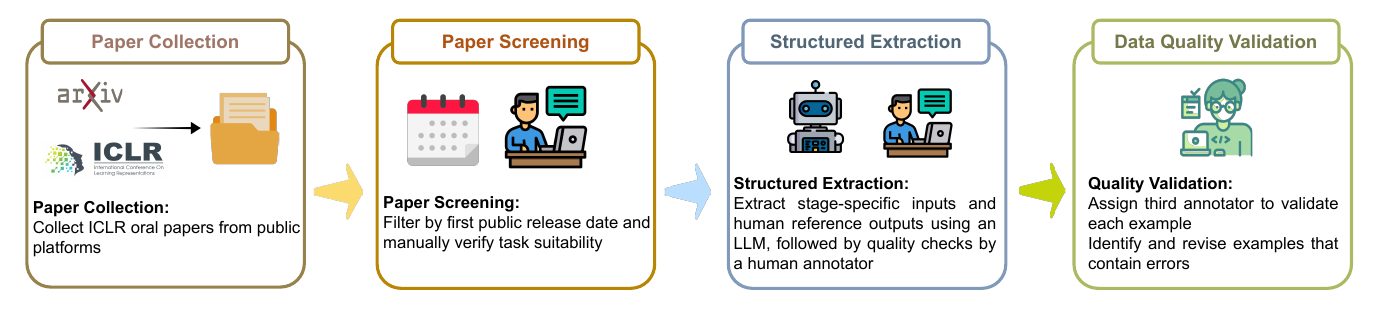}

    \caption{\textbf{Overview of evaluation corpus construction.}}

    \label{figure:corpus_overview}
\end{figure*}

\mypara{Findings-Grounded Paper Synthesis} 
The final stage of the research process is to synthesize experimental results and findings into a research paper. 
In this task, the auto researcher is provided with the research idea $I$, the experiment plan $P$, and the resulting experimental evidence and findings, denoted as $F$. 
Given these inputs, the auto researcher is tasked with synthesizing them into a coherent research paper, allowing us to assess whether it can organize, interpret, and communicate research findings in a manner comparable to human researchers.

The input for each stage is independently extracted from the corresponding target paper rather than generated from the output of the preceding stage. 
This design allows each stage to be evaluated independently without propagating errors from earlier stages. 
Details on how task inputs and corresponding human reference outputs are extracted and constructed are provided in Appendix~\ref{appendix:Stage}.

\subsection{Evaluation Corpus}
\label{section:corpus}

We construct the evaluation corpus from ICLR 2026 oral papers, using each paper as the source for both stage-specific task inputs and the corresponding human reference outputs. 
To prevent auto researchers from directly retrieving information about the ground-truth papers, we further restrict the corpus according to the knowledge cutoff of the underlying models. 
We also manually screen the remaining papers and remove those that do not satisfy the requirements of our evaluation tasks, such as papers without publicly available reference papers and purely theoretical papers without any experiments.
Following the stage-wise task definitions in Section~\ref{section:stage-wise}, we then derive the corresponding evaluation instances from each paper, pairing the stage-specific input with the human research output used as the reference for evaluation. 
All stage-specific inputs and reference outputs are extracted using an LLM-based pipeline and reformulated into structured representations aligned with the input and output formats of the respective tasks. 
Each resulting instance is reviewed and validated by human annotators to ensure the correctness and quality of the extracted content. 
After filtering and human validation, the resulting corpus contains 152 papers and 456 research trajectories.

\subsection{Checklist-based Evaluation}
\label{section:judge}

At each research stage, AutoHRS takes an auto-researcher-generated artifact and the corresponding human reference output as input. 
It evaluates the generated artifact using a human-verified checklist derived from the reference and produces item-level judgments, dimension-level scores, and an overall score measuring reconstruction fidelity.

\mypara{Checklist Evaluation Pipeline} 
We design a checklist-based evaluation protocol that enables reliable and reproducible automated assessment. 
For each stage, we first extract checklist questions from the reference output using an LLM generator. 
Each checklist question follows a general form, ``Does the output $\mathcal{O}$ contain or reflect the element $\mathcal{E}$?'', where $\mathcal{O} \in \{\text{idea}, \text{experiment plan}, \text{paper}\}$ denotes the auto researcher output at the corresponding stage, and $\mathcal{E}$ denotes an element identified in the reference output. 
An LLM then serves as a judge to evaluate the outputs generated by the auto researcher, assigning each checklist item one of three labels: \textit{Yes}, \textit{Partial}, or \textit{No}. During preliminary evaluation, we found that some checklist questions generated by the LLM were overly generic or insufficiently discriminative. 
To improve checklist quality, we introduce a human verification step in which annotators review the generated questions and remove or revise those that are ambiguous, overly broad, or uninformative for reconstruction evaluation. 
The resulting checklist is then used to assess how well the auto researcher reconstructs the corresponding human research output at each stage. 
Further details on checklist generation are provided in Appendix~\ref{appendix:checklist}.

\mypara{Stage-Specific Evaluation Dimensions}
Each stage of scientific research involves multiple interconnected decisions and outputs, making it difficult to characterize reconstruction fidelity with a single holistic criterion. 
To enable a more fine-grained evaluation, we decompose each research stage into a set of stage-specific evaluation dimensions.
Our dimension design is informed by established conventions in scientific reasoning and reporting, including the CARS framework for framing research problems and gaps, principles of rigorous and reproducible experimental design, and the IMRaD structure for scientific reporting~\cite{S90,SP04}. 
Specifically, (1) \textbf{Idea Generation}: we evaluate the generated artifact along the dimensions of \textit{Problem}, \textit{Research Gap}, \textit{Motivation}, and \textit{Method}.
(2) \textbf{Experiment Design}: we consider \textit{Experimental Objective}, \textit{Experimental Setup}, \textit{Baselines and Comparisons}, \textit{Evaluation Protocol}, and \textit{Ablation and Analysis}. 
(3) \textbf{Paper Synthesis}: we evaluate \textit{Research Framing}, \textit{Content Selection and Emphasis}, \textit{Scientific Argumentation}, \textit{Information Organization}, and \textit{Interpretation and Positioning}. 
Detailed definitions and the corresponding examples for each dimension are provided in Appendix~\ref{appendix:dimension}.

\begin{table*}[!t]
\centering
\caption{Instance-level Spearman's correlation ($\rho$) with human judgments across the three stages. Higher values indicate stronger alignment with human annotators.}
\label{table:human_alignment_idea}

\begin{tabular}{lccc}
\toprule
\textbf{Evaluator}
& \textbf{Idea Generation}
& \textbf{Experiment Design}
& \textbf{Paper Synthesis} \\
\midrule
GPT-6 Astra
& 0.60 & 0.57 & 0.64 \\

Claude Fable 5.1
& 0.62 & 0.59 & 0.60 \\

Qwen3.8-2.4T-A95B
& 0.58 & 0.50 & 0.55 \\

\midrule
\textbf{Checklist Metrics (Ours)}
& \textbf{0.70} & \textbf{0.64} & \textbf{0.68} \\
\bottomrule
\end{tabular}
\end{table*}

\mypara{Validation Against Human Judgments}
To validate whether the checklist-based scores align with human assessment, we randomly sample 20 examples from the evaluation corpus and ask human annotators to independently assess how well each auto-researcher output reconstructs the corresponding human reference. 
We further compare our checklist-based metric with a single-LLM-as-a-judge baseline. 
As shown in Table~\ref{table:human_alignment_idea}, AutoHRS achieves instance-level Spearman's correlations of $\rho=0.70$, $0.64$, and $0.68$ with human judgments for idea generation, experiment design, and paper synthesis, respectively, achieving higher observed correlations than the single-LLM judges on the annotated subset. 
These results indicate that the structured checklist provides a more reliable proxy for human evaluation. 
Annotator biographies, annotation details, and their responsibilities in the human evaluation process, together with details of the single-LLM-as-a-judge baseline, are provided in Appendix~\ref{appendix:bio},~\ref{appendix:single}, and~\ref{appendix:human}.


\section{Experiments}
In our experiments, we leverage AutoHRS to evaluate auto researchers using our curated human research trajectories as references. 
We then analyze where and how auto researchers diverge from human research, characterizing recurring patterns across different stages of the research process. 
Finally, we investigate factors that can make auto-researcher outputs more similar to human research, including the granularity of the input research direction and variation across different research types.

\subsection{Setup}

\mypara{Auto Researchers} 
We conduct our main experiments with Claude Code, using Claude Sonnet 4.5 as the underlying model. 
We select this model because its knowledge cutoff predates the target papers in our evaluation corpus, reducing the risk that the model can recover ground-truth research content from prior knowledge rather than from the provided inputs. 
To assess robustness across agent harnesses, we further replicate a subset of the experiments with Gemini CLI and Codex; setup details and additional results are provided in Appendix~\ref{appendix:auto_researcher_config} and~\ref{appendix:harness}.

\mypara{Data}
We conduct our experiments on the evaluation corpus introduced in Section~\ref{section:corpus}, constructed from ICLR 2026 oral papers. 
The corpus is filtered according to the knowledge cutoff of the underlying models and manually screened to ensure that each paper supports all three evaluation tasks. 
The final corpus contains 152 papers, yielding 456 research trajectories across \textit{idea generation}, \textit{experiment design}, and \textit{paper synthesis}. 
Each trajectory consists of the stage-specific input provided to the auto researcher and the corresponding human reference output used for evaluation.

\mypara{Metrics}
We use a unified metric formulation to evaluate auto researchers across the three research stages considered in AutoHRS. 
At each stage, the generated research artifact is evaluated along multiple dimensions, as described in Section~\ref{section:judge}. 
For each evaluation dimension $d$, the checklist-based evaluator assigns a score $s_{d,i} \in \{0, 0.5, 1\}$ to the $i$-th checklist item, corresponding to \textit{No}, \textit{Partial}, and \textit{Yes}, respectively. 
Since the number of checklist items may vary across dimensions, we first compute a dimension-level score by averaging over all checklist items within that dimension:
\begin{equation}
\label{equ:stage_score}
S_d = \frac{1}{N_d}\sum_{i=1}^{N_d} s_{d,i}
\end{equation}
where $N_d$ denotes the number of checklist items associated with dimension $d$. 
To aggregate dimension-level scores into an overall AutoHRS score, we assign stage-specific weights to the evaluation dimensions. 
The weights are selected based on alignment with human judgments, using the weighting scheme that yields the highest correlation with human evaluation for each research stage. 
Further details on the weight selection procedure and the resulting stage-specific weights are provided in Appendix~\ref{appendix:weight}. 
For a given stage $t$, the overall AutoHRS score is computed as a weighted average of the corresponding dimension-level scores:
\begin{equation}
\label{equ:autoh​rs_score}
S_{\mathrm{AutoHRS}}^{(t)}
=
\sum_{d=1}^{D_t} w_d^{(t)} S_d^{(t)},
\qquad
\sum_{d=1}^{D_t} w_d^{(t)} = 1
\end{equation}
where $w_d^{(t)}$ denotes the weight assigned to dimension $d$ at stage $t$, and $D_t$ denotes the number of evaluation dimensions defined for that stage. 
Finally, we report $S_{\mathrm{AutoHRS}}^{(t)}$ separately for each research stage. 
Higher AutoHRS scores indicate greater fidelity in reconstructing the corresponding human research output and stronger alignment with the human reference.

\subsection{Main Findings}

\begin{figure*}[!t]
    \centering
    \includegraphics[width=0.90\textwidth]{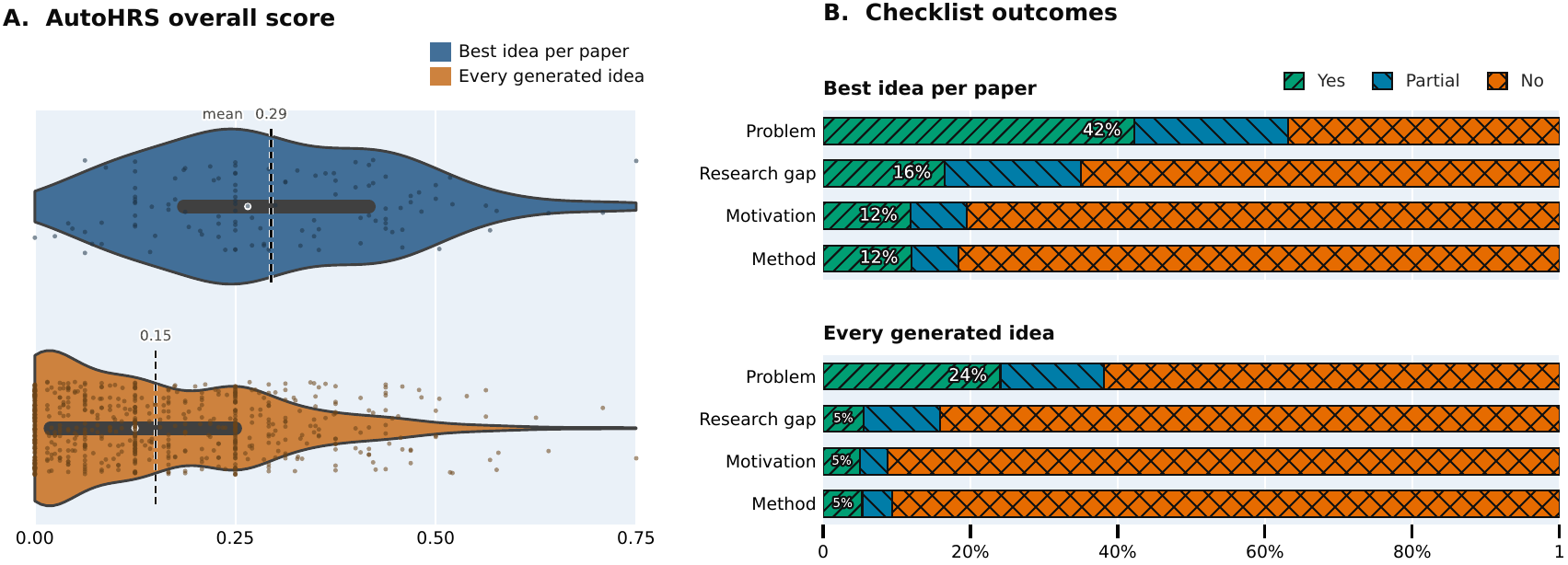}
    \caption{\textbf{Idea generation results.}
    (A) Distributions of overall AutoHRS scores for every generated idea and the best-scoring idea for each paper.
    ``\meanline~'' indicates the mean and
    ``\medianpoint~'' indicates the median.
    (B) Proportions of checklist outcomes (\textit{Yes}, \textit{Partial}, and \textit{No}) within each evaluation dimension, shown separately for the best idea per paper and every generated idea.}
    \label{figure:idea_generation}
\end{figure*}
\begin{figure*}[tb]
    \centering
    \includegraphics[width=0.90\textwidth]{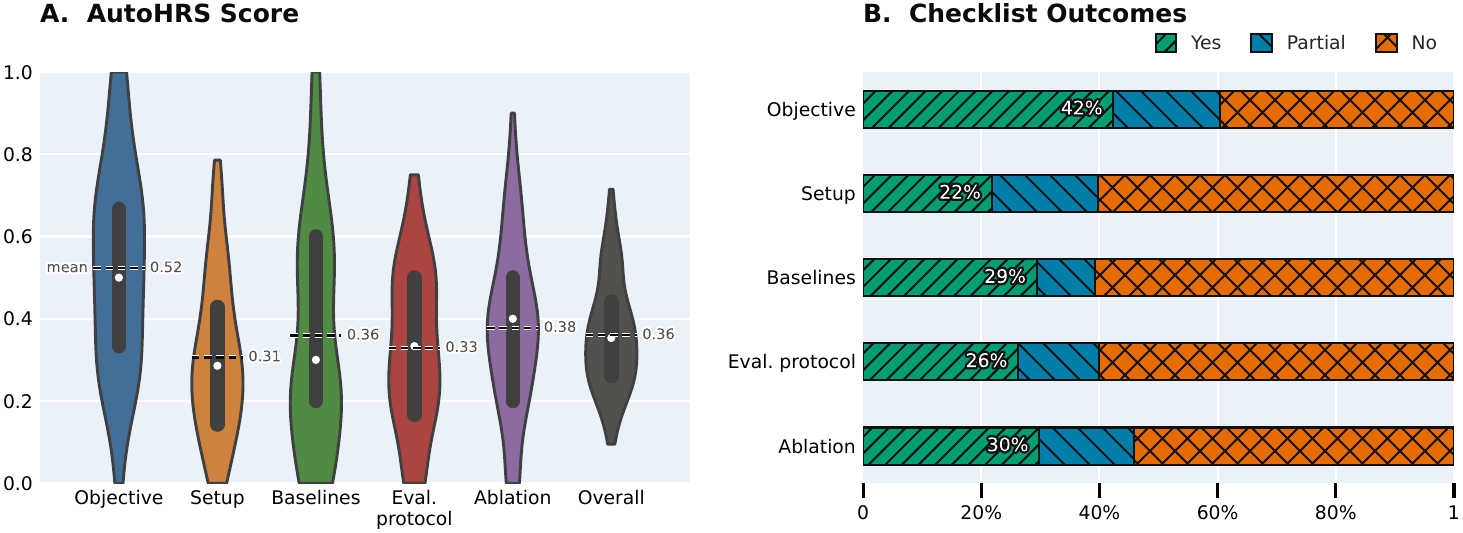}
    \caption{\textbf{Experiment design results.}
    (A) Distributions of AutoHRS scores for each evaluation dimension and the overall score.
    (B) Proportions of checklist outcomes (\textit{Yes}, \textit{Partial}, and \textit{No}) within each evaluation dimension.}
    \label{figure:experiment_design}
\end{figure*}

\mypara{Idea Generation}
Figure~\ref{figure:idea_generation} shows that auto researchers achieve limited alignment with human references in literature-grounded idea generation. 
The mean overall AutoHRS score is 0.29 for the best-scoring idea per paper and 0.15 across all generated ideas. 
Regarding the proportions of checklist outcomes, the results show that, for the best idea per paper, 42\% of problem-related checklist items receive a \textit{Yes}, whereas the corresponding proportions are 16\% for research gap identification and 12\% for both motivation and method. 
Across all generated ideas, 24\% of problem-related items receive a \textit{Yes}, while this proportion is only 5\% for each of the remaining dimensions. 
In both groups, \textit{No} outcomes dominate the research gap, motivation, and method dimensions. 
These results indicate that auto researchers are better at recovering the broad research problem than reconstructing the more specific research decisions underlying the human reference.

\mypara{Experiment Design}
Figure~\ref{figure:experiment_design} shows that auto researchers achieve a mean overall AutoHRS score of 0.36 in idea-grounded experiment design. 
Alignment is strongest in \textit{Experimental Objective}, with a mean score of 0.52 and 42\% of checklist items receiving a \textit{Yes}. 
The remaining dimensions exhibit lower alignment: mean scores are 0.31 for \textit{Experimental Setup}, 0.36 for \textit{Baselines and Comparisons}, 0.33 for \textit{Evaluation Protocol}, and 0.38 for \textit{Ablation and Analysis}, with corresponding \textit{Yes} proportions of 22\%, 29\%, 26\%, and 30\%, respectively.
\textit{No} outcomes account for more than half of the checklist items in each of these four dimensions. 
Although the overall AutoHRS score is higher than in idea generation, alignment with human references remains poor in concrete experimental choices, including experimental setup, baseline selection, evaluation protocols, and ablation studies.

\begin{figure*}[!t]
    \centering
    \includegraphics[width=0.95\textwidth]{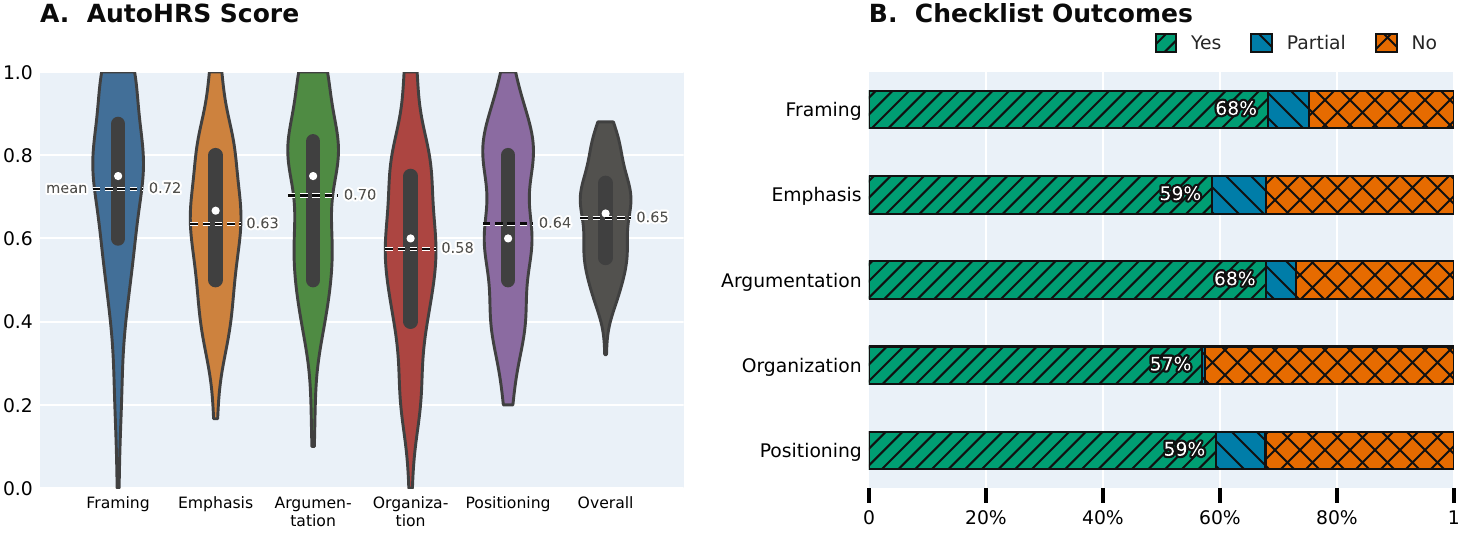}
    \caption{\textbf{Paper synthesis results.}
    (A) Distributions of AutoHRS scores for each evaluation dimension and the overall score.
    (B) Proportions of checklist outcomes (\textit{Yes}, \textit{Partial}, and \textit{No}) within each evaluation dimension.}
    \label{figure:Paper_Synthesis}
\end{figure*}

\mypara{Paper Synthesis}
Figure~\ref{figure:Paper_Synthesis} shows that auto researchers achieve a mean overall AutoHRS score of 0.65 in findings-grounded paper synthesis. 
Alignment is strongest in \textit{Research Framing} and \textit{Scientific Argumentation}, with mean scores of 0.72 and 0.70, respectively, and 68\% of checklist items receiving a \textit{Yes} in both dimensions. 
Mean scores are 0.63 for \textit{Content Selection and Emphasis} and 0.64 for \textit{Interpretation and Positioning}, with a \textit{Yes} proportion of 59\% in each dimension. 
\textit{Information Organization} exhibits the lowest alignment, with a mean score of 0.58 and 57\% of checklist items receiving a \textit{Yes}. 
These results suggest that auto researchers are more effective at reconstructing human research in paper writing than in earlier stages when provided with the research context and findings. 
However, auto researchers still differ from human experts in how they organize papers and emphasize key aspects of the research.

\begin{tcolorbox}[
title=Takeaway,
colback=cyan!3,
colframe=cyan!45!black,
fontupper=\small,
fonttitle=\bfseries\small,
boxrule=0.6pt,
arc=1mm,
before skip=8pt,
after skip=8pt]
AutoHRS scores are 0.29 for the best-scoring idea generation, 0.36 for experiment design, and 0.65 for paper synthesis. Despite these differences across stages, substantial gaps from human research remain in all three, suggesting that current auto researchers still struggle to reproduce the research decisions and outputs of human researchers.
\end{tcolorbox}

\subsection{Where and how auto researchers diverge from human research}

\mypara{Reference-Driven Idea Composition} 
During manual inspection of auto-researcher-generated research trajectories, we observed several recurring patterns that may contribute to their divergence from human-generated ideas.
Auto researchers frequently construct new ideas by combining or integrating elements from multiple reference papers, whereas such patterns appear less frequently in the corresponding human ideas. 
To examine this observation, we randomly sample 20 research trajectories and manually assess whether each idea is primarily formed by combining identifiable elements from the provided reference papers. 
Under this criterion, 17 of the 20 auto-researcher-generated ideas exhibit a composition-based pattern, whereas the corresponding human ideas do not. 
We then extend this analysis to all research trajectories using \textit{GPT-6 Astra} to assess the relationship between generated ideas and their reference papers. 
The results show that 78.4\% of auto-researcher-generated ideas exhibit this pattern, suggesting a strong tendency to synthesize existing reference elements rather than depart from them in the same way as human researchers.

\mypara{Conventional Biases in Experiment Planning}
Recurring patterns are also observed in auto-researcher-generated experiment plans. 
Rather than tailoring experimental designs closely to the specific research question, auto researchers often fall back on conventional experimental choices. 
They tend to select commonly used models and standard evaluation metrics, while introducing task-specific or customized metrics much less frequently. 
Together, these patterns suggest that auto researchers often rely on generic experimental templates, which may contribute to their divergence from human-designed experiment plans.

\begin{tcolorbox}[
title=Takeaway,
colback=cyan!3,
colframe=cyan!45!black,
fontupper=\small,
fonttitle=\bfseries\small,
boxrule=0.6pt,
arc=1mm,
before skip=8pt,
after skip=8pt]
Auto researchers exhibit recurring patterns in both idea generation and experiment planning: 78.4\% of generated ideas recombine elements from the provided references, while experiment plans often favor common models and standard metrics over more tailored designs.
\end{tcolorbox}

\subsection{How to Make Auto-Researcher Outputs More Similar to Human Research}

We further investigate factors that may affect the reconstruction performance of auto researchers. 
We examine how the granularity of the input research direction influences idea generation, and whether reconstruction performance varies across different types of research papers, including \textit{Empirical}, \textit{Methodological}, and \textit{Benchmark-oriented} work.

\begin{figure}[t!]
    \centering
    \includegraphics[width=0.95\columnwidth]{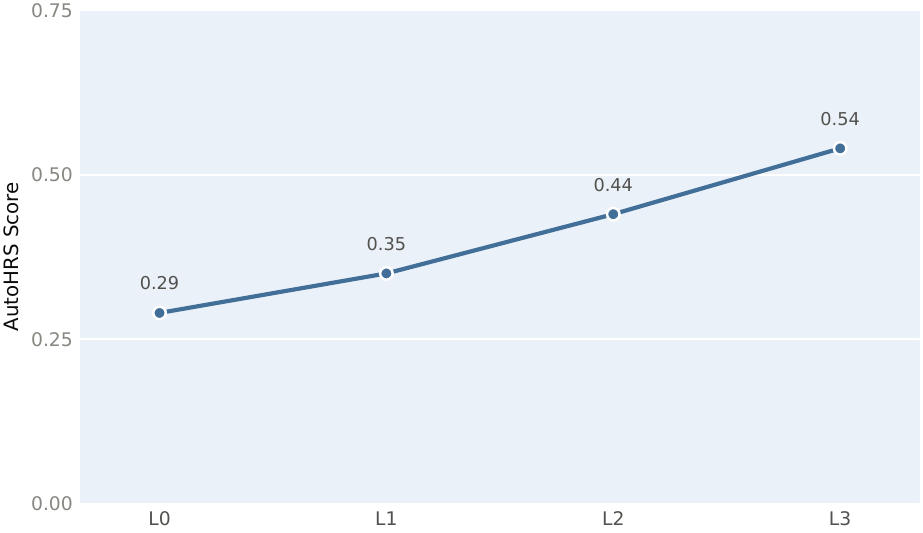}
    \caption{\textbf{Impact of research-direction granularity on idea-generation performance.}
    Overall AutoHRS score at different levels of research-direction granularity (L0--L3).}
    \label{figure:granularity}
\end{figure}

\mypara{Impact of Research Direction Granularity}
In our main experiments, the input to the idea generation stage consists of reference papers together with a research direction.
By default, the research direction is expressed as a broad phrase, such as ``LLM Reasoning'', ``Multimodal Learning'', or ``Agent Safety''.
In preliminary experiments, we observed that providing slightly more specific research directions can improve idea-generation performance. 
Motivated by this observation, we investigate how the granularity of the research direction affects reconstruction performance.
We define four levels of granularity by progressively narrowing the topical scope while avoiding information that directly reveals the target problem, research gap, motivation, or method. 
The four levels correspond to a \textit{broad field}, a \textit{research topic}, a \textit{focused research area}, and a \textit{fine-grained research scope}. 
Examples for each granularity level are provided in Appendix~\ref{appendix:gra}. 
Each level introduces only a modest amount of additional information over the previous one.
As shown in Figure~\ref{figure:granularity}, AutoHRS scores consistently increase as the research direction becomes more fine-grained.

\begin{figure}[tb]
    \centering
    \includegraphics[width=0.95\columnwidth]{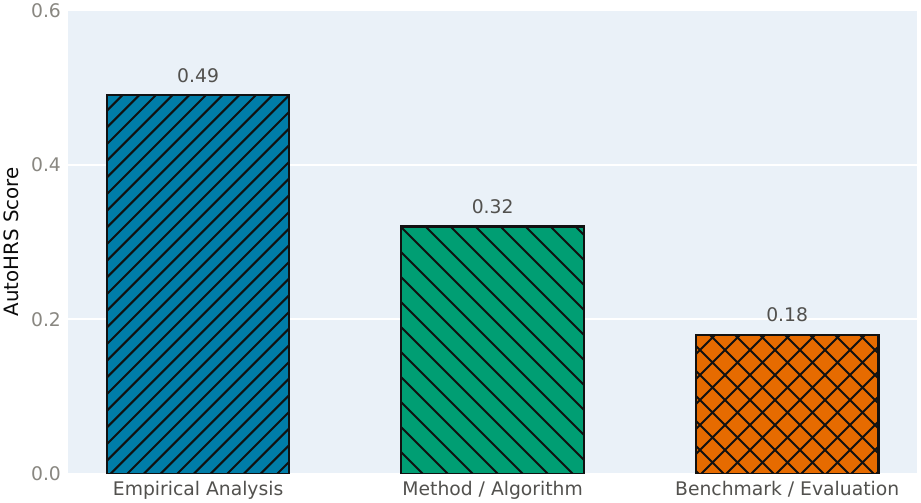}
    \caption{\textbf{Impact of research type on idea-generation performance.}
    Overall AutoHRS score across different research types: \textit{Empirical Analysis}, \textit{Method/Algorithm}, and \textit{Benchmark/Evaluation}.}
    \label{figure:research_type}
\end{figure}

\mypara{Performance Across Research Types}
To examine whether reconstruction performance varies across research types, we manually identify papers in the evaluation corpus that fall into three categories: \textit{Empirical Analysis}, \textit{Method/Algorithm}, and \textit{Benchmark/Evaluation}.
For each category, we select 20 papers that match the corresponding research type and evaluate idea-generation performance within each subset. 
As shown in Figure~\ref{figure:research_type}, auto researchers achieve the highest AutoHRS score of 0.49 on \textit{Empirical Analysis} papers, followed by 0.32 on \textit{Method/Algorithm} papers and 0.18 on \textit{Benchmark/Evaluation} papers. 
These results suggest that the ability of auto researchers to reconstruct human research ideas varies substantially across research types, with benchmark- and evaluation-oriented work being particularly challenging.

\begin{tcolorbox}[
title=Takeaway,
colback=cyan!3,
colframe=cyan!45!black,
fontupper=\small,
fonttitle=\bfseries\small,
boxrule=0.6pt,
arc=1mm,
before skip=8pt,
after skip=8pt]
More fine-grained research directions consistently improve idea reconstruction. Performance also varies by research type, with AutoHRS scores of 0.49 for empirical analysis, 0.32 for methodological work, and 0.18 for benchmark and evaluation research.
\end{tcolorbox}

\section{Related Work}
\label{sec:related work}

\subsection{Automated Scientific Research}

Recent advances in large language models and general-purpose agent harnesses have substantially broadened the scope of automated scientific research. 
Early LLM-based work largely focused on individual stages of the research process, such as idea generation, experimentation, and scientific writing~\cite{HFWWLXLJPL24, BJCH25, GDWCCW24, SSWSWYLMLB25, JSSXKJW25, CSV24, CLWHZWWH26, YWLYLZCD26}. 
More recently, harness-based auto researchers have begun to operate across the full research workflow, shifting the role of humans increasingly toward high-level guidance and oversight. 
ARIS~\cite{YLL26} enables end-to-end autonomous research through adversarial multi-agent collaboration, while Sibyl-AutoResearch~\cite{WXHGWX26}, a self-evolving research harness, improves through trial-and-error experience. 
In this work, we focus specifically on harness-based auto researchers that carry out multiple stages of the scientific research process in an end-to-end manner.

\subsection{Evaluation of Auto Researchers}

Existing evaluations of automated scientific research largely focus either on model-level capabilities or on task completion within individual stages of the research process~\cite{CCNZWYLLWLDXBBAHNGSS25, ZWGC26}. 
RE-Bench~\cite{WLBJPBCCCDEGGJKLNSSTWB25} compares language-model agents with human experts on open-ended machine-learning research engineering tasks. 
IdeaBench~\cite{GSXHKWBZ25} evaluates LLMs on research idea generation, focusing on the quality of generated ideas. 
In contrast, the systematic evaluation of general-purpose harness-based auto researchers has only recently begun to emerge. 
The work by~\cite{FLYCLTCSMD26} focuses on diagnosing and tracing the causes of failures in end-to-end AutoResearch trajectories. 
A concurrent work, ARAC~\cite{CYZXWL26}, evaluates auto-research systems against generalized human research methodology, whereas we directly assess whether they reconstruct the specific research decisions of individual human research trajectories.

\section{Conclusion and Discussion}

We introduced AutoHRS, a stage-wise framework for evaluating how faithfully auto researchers reconstruct human research across idea generation, experiment design, and paper synthesis. 
AutoHRS further provides a fine-grained checklist-based evaluation pipeline with stage-specific dimensions, and we validate its alignment with human judgments. 
Our results reveal substantial differences from human research across all three stages, with larger gaps in earlier research decisions. 
Further analysis shows recurring patterns in how auto researchers diverge from human researchers, including reference-driven idea composition and conventional experiment planning.
We also find that reconstruction performance can improve with more fine-grained research directions and varies across research types.
These findings suggest that evaluating auto researchers requires looking beyond end-to-end task completion to the research decisions and processes underlying their outputs. 
Future auto-researcher systems should therefore aim not only to automate scientific workflows, but also to make research decisions that more closely resemble those of human researchers.

\section{Limitations}

Our study has several limitations.
First, AutoHRS is evaluated only on research papers from the AI domain, and it remains unclear whether the observed patterns generalize to other scientific disciplines with different research practices and evaluation criteria. 
Second, running end-to-end auto-researcher trajectories is computationally expensive. 
As a result, our main evaluation is limited to 152 papers, corresponding to 456 stage-wise research trajectories. 
While this scale is sufficient to reveal consistent patterns across our analyses, broader evaluations over larger and more diverse corpora would provide a more comprehensive assessment of auto-researcher behavior.

\bibliographystyle{plain}
\bibliography{normal_generated_py3}

\newpage
\appendix

\section{AutoHRS Construction}

\subsection{Biography of Annotators}
\label{appendix:bio}

\begin{table}[!ht]
\centering
\caption{Details of the human annotators involved in checklist verification and human evaluation. All annotators have research experience in AI and independently assess the reconstruction fidelity of auto-researcher outputs against the corresponding human references.}
\label{tab:candidate_profiles}

\resizebox{\columnwidth}{!}{%
\begin{tabular}{lllcccc}
\toprule
\textbf{ID}
& \textbf{Domain}
& \textbf{\# Publications}
& \textbf{Prompt}
& \textbf{Checklist}
& \textbf{Validation}
& \textbf{Human Eval} \\
\midrule
1 & Machine Learning & $>$ 10 & \cmark & \cmark & \cmark & \cmark \\
2 & Natural Language Processing & 5--10 & \cmark & \cmark & \cmark & \cmark \\
3 & Computer Vision & 5--10 & \cmark & \cmark & \cmark & \cmark \\
\bottomrule
\end{tabular}%
}
\end{table}


\subsection{Stage-Wise Evaluation Task Construction}
\label{appendix:Stage}

This section provides additional details on the construction of the stage-wise evaluation tasks. We use GPT-6 Astra to extract and reformulate the task inputs and corresponding human reference outputs from the target papers. Because the extraction prompts for the three research stages follow a similar structure, we present the research-idea extraction prompt as a representative example below.

\begin{figure}[htbp]
\centering
\begin{tcolorbox}[
width=\linewidth,
colback=black!7.5!white,
colframe=black!80!white,
title=Prompt for Research Idea Extraction,
fontupper=\scriptsize,
fonttitle=\scriptsize,
boxrule=0.4pt,
sharp corners,
before upper={\linespread{0.92}\selectfont},
]
\texttt{[User Input]}
\par\smallskip
You are an expert ML/NLP research reviewer. Given a research paper,
extract the core research idea proposed by the authors and represent it
in a structured form. The extracted idea should capture the research
decision made before experiment design, rather than details that can
only be inferred from the experimental results.

\par\smallskip
\textbf{Extraction requirements.}
\begin{itemize}
\setlength{\itemsep}{0pt}
\setlength{\parskip}{0pt}

\item \textbf{Research problem:} Identify the concrete problem or
task that the paper aims to address.

\item \textbf{Research gap:} Identify the limitation or unresolved
issue in existing work that motivates the proposed research.

\item \textbf{Motivation:} Extract the key observation, intuition,
or reasoning that motivates the proposed approach.

\item \textbf{Method:} Describe the proposed method and its core
technical mechanism, including the design choices that distinguish
it from existing approaches.

\end{itemize}

\textbf{General requirements.}
\begin{itemize}
\setlength{\itemsep}{0pt}
\setlength{\parskip}{0pt}

\item Use only information supported by the provided paper.
Do not introduce external knowledge or infer unsupported claims.

\item Focus on the research idea itself rather than implementation
details, experimental settings, quantitative results, or conclusions
derived from the experiments.

\item Preserve the substantive technical content of the paper while
reformulating it into a concise and self-contained representation.

\item Separate the research motivation from the
proposed method so that the resulting idea can be used independently
as input for experiment planning.

\end{itemize}

\textbf{Inputs:}\
Paper title: {title}\
Paper content: {paper}

\par\smallskip
\textbf{Output format:}\
Return only a JSON object without markdown fences.

\par\smallskip
{\ttfamily
\noindent
\{\
\hspace*{1em}"problem": "<research~problem>",\
\hspace*{1em}"research\_gap": "<research gap>",\
\hspace*{1em}"motivation": "<research motivation>",\
\hspace*{1em}"method": "<proposed method>"\
\}
\par
}

\end{tcolorbox}
\caption{
Prompt for extracting the structured research idea from the reference paper.
The extracted idea captures the research problem, gap, motivation, and proposed method, and serves as the input to the experiment design stage.
}
\label{figure:prompt_idea_extraction}
\end{figure}

\subsection{Checklist Generation}
\label{appendix:checklist}

This section provides the prompts used in our checklist-based evaluation pipeline. We first present the stage-specific prompts used to generate checklist questions for \textit{idea generation}, \textit{experiment design}, and \textit{paper synthesis}. We then provide the prompt used by the LLM judge to evaluate auto-researcher outputs against each checklist item and assign one of three outcomes: \textit{Yes}, \textit{Partial}, or \textit{No}.
\begin{figure}[htbp]
\centering
\begin{tcolorbox}[
  width=\linewidth,
  colback=black!7.5!white,
  colframe=black!80!white,
  title=Prompt for Idea Checklist Generation,
  fontupper=\scriptsize,
  fonttitle=\scriptsize,
  boxrule=0.4pt,
  sharp corners,
  before upper={\linespread{0.92}\selectfont},
]

\texttt{[User Input]}
\par\smallskip
You are an expert ML/NLP research reviewer. Decompose the reference idea
into a checklist of atomic YES/NO questions for assessing whether a
candidate idea contains the same research elements.

\par\smallskip
\textbf{Dimensions and question limits.}
Use the following dimensions in this order:
\begin{enumerate}
    \setlength{\itemsep}{0pt}
    \setlength{\parskip}{0pt}
    \item \texttt{problem} (1--5): The concrete research problem or task setting.
    \item \texttt{research\_gap} (1--5): Limitations of existing work that
    leave the problem unresolved, rather than a restatement of the problem.
    \item \texttt{motivation} (1--5): The observation, insight, or hypothesis
    explaining why this approach should work, rather than why existing
    work is inadequate.
    \item \texttt{method} (1--8): The core technical mechanism and
    distinctive design details.
\end{enumerate}
Generate as many questions as the reference supports, with at most
20 questions overall. Do not pad dimensions or combine distinct
commitments to meet these limits.

\par\smallskip
\textbf{Question requirements.}
\begin{itemize}
    \setlength{\itemsep}{0pt}
    \setlength{\parskip}{0pt}
    \item \textbf{Atomic:} Test exactly one element per question.
    Split independently testable claims; use conjunctions only for
    indivisible concepts.
    \item \textbf{Grounded:} Use only content stated in the reference idea.
    Include an exact supporting substring as \texttt{source\_span};
    do not import external knowledge or invent content.
    \item \textbf{Concrete:} Ask about substantive content, not generic
    similarity or method names alone. For example, ask whether the idea
    marks a target pixel visually, rather than whether it uses the
    ``same method.''
    \item \textbf{Self-contained:} Each question must be answerable from
    the candidate idea alone. Avoid citations and references to
    ``the paper,'' ``the original idea,'' or ``above.''
    \item \textbf{Distinct:} Avoid duplicates within and across dimensions.
    Keep the problem, limitations of existing work, and rationale for
    the proposed approach separate.
    \item \textbf{Match-oriented:} YES must indicate a match.
    Avoid negations. Accept equivalent wording while rejecting
    ideas that merely share the same subfield.
    \item \textbf{Coverage:} Include at least one question per dimension.
    If support is sparse, use the best-supported question and leave
    \texttt{source\_span} empty rather than inventing content.
\end{itemize}

\textbf{Inputs:}\\
Research direction: \{direction\}\\
Paper title: \{title\}\\
Reference idea: \{idea\}

\par\smallskip
\textbf{Output format:}\\
Briefly reason about the distinct commitments in each dimension,
then return a JSON object without markdown fences.
Order the items by \texttt{problem}, \texttt{research\_gap},
\texttt{motivation}, and \texttt{method}.

\par\smallskip
{\ttfamily
\noindent
\{"checklist": [\\
\hspace*{1em}\{"dimension": "<dimension>",\\
\hspace*{2em}"element": "<single element as a noun phrase>",\\
\hspace*{2em}"source\_span": "<exact reference substring>",\\
\hspace*{2em}"question": "<self-contained YES/NO question>"\},\\
\hspace*{1em}...\\
]\}
\par
}

\end{tcolorbox}
\caption{
Prompt for checklist generation in the idea generation stage.
The reference idea is decomposed into atomic questions across
problem, research gap, motivation, and method.
}
\label{figure:prompt_checklist_generation_idea}
\end{figure}

\begin{figure}[htbp]
\centering
\begin{tcolorbox}[
  width=\linewidth,
  colback=black!7.5!white,
  colframe=black!80!white,
  title=Prompt for Experiment Design Checklist Generation,
  fontupper=\scriptsize,
  fonttitle=\scriptsize,
  boxrule=0.4pt,
  sharp corners,
  before upper={\linespread{0.92}\selectfont},
]

\texttt{[User Input]}
\par\smallskip
You are an expert ML/NLP research reviewer. Decompose the reference
experiment plan into atomic YES/NO questions for assessing whether
a candidate plan contains the same experimental decisions.

\par\smallskip
\textbf{Dimensions and question limits.}
Use the following dimensions in this order:
\begin{enumerate}
    \setlength{\itemsep}{0pt}
    \setlength{\parskip}{0pt}
    \item \texttt{experimental\_objective} (1--4):
    The claims the experiments aim to establish.
    \item \texttt{experimental\_setup} (1--7):
    Datasets, benchmarks, systems, backbones, and implementation
    decisions such as training budgets, decoding, and environments.
    \item \texttt{baselines\_and\_comparisons} (1--5):
    Prior methods, comparison variants, human or oracle references,
    and conditions ensuring fair comparisons.
    \item \texttt{evaluation\_protocol} (1--6):
    Metrics, evaluation splits, repeated runs, aggregation,
    human-study design, judges, verifiers, and uncertainty reporting.
    \item \texttt{ablation\_and\_analysis} (1--5):
    Component ablations, parameter or scale sweeps,
    diagnostic breakdowns, and qualitative analyses.
\end{enumerate}
Generate only as many questions as the reference supports, with at most
27 overall. Within each dimension, order questions by centrality and
retain the most central commitments if the cap is exceeded.
Do not pad dimensions or merge independent commitments to fit the caps.
Group benchmark items only when they express one evaluation-data
commitment, leaving room for other setup decisions.

\par\smallskip
\textbf{Question requirements.}
\begin{itemize}
    \setlength{\itemsep}{0pt}
    \setlength{\parskip}{0pt}
    \item \textbf{Plans, not results:} Ask what the candidate would run,
    compare, measure, or vary, never what it would find or what numerical
    results it would obtain. Ask whether it ablates a memory module,
    not whether removing that module reduces accuracy by a given amount.
    \item \textbf{Atomic:} Test exactly one element per question.
    Split independently testable claims; conjunctions may describe
    only an indivisible concept.
    \item \textbf{Grounded:} Use only the reference plan.
    Cite supporting item IDs in \texttt{source\_items}.
    Multiple IDs are allowed when they support one commitment.
    Do not invent content or import external knowledge.
    \item \textbf{Concrete:} Test substantive experimental decisions,
    not generic appropriateness, method names alone, or arbitrary
    constants. For example, ask about training on preference pairs,
    not matching an exact learning rate.
    \item \textbf{Self-contained:} Each question must be answerable
    from the candidate plan alone. Avoid citations within questions
    and phrases such as ``the reference plan'' or ``as described above.''
    \item \textbf{Distinct:} Avoid duplicate questions within and
    across dimensions. Put baselines under comparisons, metrics
    under evaluation protocol, and ablations under ablation and analysis.
    \item \textbf{Match-oriented:} YES must indicate a match.
    Avoid negations; accept equivalent wording while rejecting plans
    that merely share the same subfield.
    \item \textbf{Coverage:} Include at least one question per dimension.
    If a dimension is empty, use the best-supported question from
    the remaining plan and leave \texttt{source\_items} empty
    rather than inventing content.
\end{itemize}

\textbf{Inputs:}\\
Paper title: \{title\}\\
Reference experiment plan: \{reference\_plan\}

\par\smallskip
\textbf{Output format:}\\
Briefly reason about the distinct commitments in each dimension,
then return a JSON object without markdown fences.
Group checklist items in the dimension order listed above.

\par\smallskip
{\ttfamily
\noindent
\{"checklist": [\\
\hspace*{1em}\{"dimension": "<dimension>",\\
\hspace*{2em}"element": "<single element as a noun phrase>",\\
\hspace*{2em}"source\_items": ["<reference item ID, e.g., SET3>"],\\
\hspace*{2em}"question": "<self-contained YES/NO question>"\},\\
\hspace*{1em}...\\
]\}
\par
}

\end{tcolorbox}
\caption{
Prompt for checklist generation in the experiment design stage.
The reference experiment plan is decomposed into atomic questions
across experimental objective, experimental setup, baselines and
comparisons, evaluation protocol, and ablation and analysis.
}
\label{figure:prompt_checklist_generation_experiment}
\end{figure}

\begin{figure}[htbp]
\centering
\begin{tcolorbox}[
  width=\linewidth,
  colback=black!7.5!white,
  colframe=black!80!white,
  title=Prompt for Paper Synthesis Checklist Generation,
  fontupper=\scriptsize,
  fonttitle=\scriptsize,
  boxrule=0.4pt,
  sharp corners,
  before upper={\linespread{0.92}\selectfont},
]

\texttt{[User Input]}
\par\smallskip
You are an expert ML/NLP research reviewer. Build a checklist of atomic
YES/NO questions assessing whether a candidate paper makes the same
writing choices as the reference paper. Both papers share the same
research content; evaluate framing, emphasis, argumentation,
organization, and interpretation rather than whether facts are mentioned.

\par\smallskip
\textbf{Dimensions and question limits.}
Use the following dimensions in this order:
\begin{enumerate}
    \setlength{\itemsep}{0pt}
    \setlength{\parskip}{0pt}
    \item \texttt{research\_framing} (1--5):
    How the problem and gap are defined, how motivation is established,
    and how background leads into the contribution.
    \item \texttt{content\_selection\_and\_emphasis} (1--6):
    Which findings, experiments, and method details receive prominence,
    and which are compressed, deferred, or treated as supporting material.
    \item \texttt{scientific\_argumentation} (1--6):
    How motivation justifies design and how evidence supports claims.
    Test the link between elements, not their mere presence.
    \item \texttt{information\_organization} (1--5):
    The relative order and grouping of content, including method
    exposition and experiment sequencing, independent of section titles.
    \item \texttt{interpretation\_and\_positioning} (1--5):
    What findings are said to mean, how contributions relate to prior
    work, which limitations are conceded, and how strongly claims are made.
\end{enumerate}
Include at least one question per dimension and at most 27 overall.
Do not pad dimensions or merge independent choices.
When a cap is exceeded, retain the most consequential choices that
another competent author could plausibly make differently.
Order questions within each dimension most-central-first.

\par\smallskip
\textbf{Question requirements.}
\begin{itemize}
    \setlength{\itemsep}{0pt}
    \setlength{\parskip}{0pt}
    \item \textbf{Discriminative:} Contrast the reference's choice with
    a plausible alternative and record it in \texttt{alternative}.
    For argumentation, specify which evidence supports which claim;
    the alternative may be another interpretation of that evidence.
    \item \textbf{Writing-focused:} Ask whether a result is presented
    as a headline finding rather than a supporting check, not merely
    whether that result appears.
    \item \textbf{Atomic:} Test exactly one writing choice per question.
    Split independently testable choices; conjunctions may describe
    only one indivisible choice.
    \item \textbf{Grounded:} Derive the reference's choice only from
    the supplied paper. Cite section IDs in \texttt{source\_sections}
    and provide a verbatim \texttt{source\_span} of at most 240 characters.
    Two excerpts may be joined with `` ... ''.
    Leave the span empty only when an ordering choice is supported
    by position rather than wording.
    \item \textbf{Self-contained:} Questions must be answerable from
    the candidate alone, without citations or references to the original.
    Describe methods and components by function, not paper-specific
    names or section titles. Shared dataset, benchmark, and prior-method
    names may be used.
    \item \textbf{Distinct:} Avoid duplicates within and across dimensions.
    Separate position, emphasis, meaning, and evidence-to-claim links.
    \item \textbf{Match-oriented:} YES must indicate the same writing
    choice. Avoid negations and accept equivalent wording, while
    rejecting choices that merely concern the same topic.
\end{itemize}

\textbf{Input:}\\
Reference paper: \{reference\_paper\}\\
Includes the title and sections labeled [S1], [S2], \ldots,
in document order.

\par\smallskip
\textbf{Output format:}\\
Briefly reason about each writing choice and its plausible alternative,
then return a JSON object without markdown fences.
Group items in the dimension order listed above.

\par\smallskip
{\ttfamily
\noindent
\{"checklist": [\\
\hspace*{1em}\{"dimension": "<dimension>",\\
\hspace*{2em}"element": "<single writing choice as a noun phrase>",\\
\hspace*{2em}"choice": "<what the reference paper did>",\\
\hspace*{2em}"alternative": "<plausible other choice>",\\
\hspace*{2em}"source\_sections": ["<section ID, e.g., S3>"],\\
\hspace*{2em}"source\_span": "<verbatim evidence, or empty string>",\\
\hspace*{2em}"question": "<self-contained YES/NO question>"\},\\
\hspace*{1em}...\\
]\}
\par
}

\end{tcolorbox}
\caption{
Prompt for checklist generation in the paper synthesis stage.
The checklist captures writing choices across research framing,
content selection and emphasis, scientific argumentation,
information organization, and interpretation and positioning.
}
\label{figure:prompt_checklist_generation_paper}
\end{figure}

\begin{figure}[htbp]
\centering
\begin{tcolorbox}[
width=\linewidth,
colback=black!7.5!white,
colframe=black!80!white,
title=Prompt for Checklist Evaluation,
fontupper=\scriptsize,
fonttitle=\scriptsize,
boxrule=0.4pt,
sharp corners,
before upper={\linespread{0.92}\selectfont},
]

\texttt{[User Input]}
\par\smallskip
You are grading whether a candidate research artifact contains one
specific element.

Read the candidate output, then answer the single checklist question
below with one verdict:

\begin{itemize}
\setlength{\itemsep}{0pt}
\setlength{\parskip}{0pt}

\item \textbf{YES} -- The candidate output contains or commits to
what the question asks. Judge the substance rather than the exact
wording; semantically equivalent content should still be considered
a match.

\item \textbf{PARTIAL} -- The candidate output covers only part of
what the question asks, expresses it only vaguely, or contains a
closely related element that falls short of the specific requirement.
Use this label when neither \textbf{YES} nor \textbf{NO} adequately
describes the match.

\item \textbf{NO} -- The candidate output does not contain the
requested element or does not mention it. Absence should be
treated as \textbf{NO}.

\end{itemize}

Judge only based on what is explicitly stated in the candidate output.
Do not infer missing information from the research area or introduce
external knowledge.

Provide a one-sentence justification of at most 25 words indicating
what the candidate output does or does not contain. Do not hedge or
simply restate the checklist question.

\par\smallskip
\textbf{Inputs:}\
Candidate research artifact: {candidate\_output}\
Checklist question: {question}

\par\smallskip
\textbf{Output format:}\
Return only a JSON object without markdown fences.

\par\smallskip
{\ttfamily
\noindent
\{"answer": "<YES, NO or PARTIAL>",\
\hspace*{1em}"reason": "<one sentence, at most 25 words>"\}
\par
}

\end{tcolorbox}

\caption{
Prompt used by the LLM judge for checklist-based evaluation.
Each checklist question is evaluated independently against the
auto-researcher-generated artifact and assigned a
\textit{Yes}, \textit{Partial}, or \textit{No} verdict together with
a brief justification.
}
\label{figure:prompt_checklist_evaluation}
\end{figure}

\subsection{Checklist Dimensions}
\label{appendix:dimension}

This section provides additional details on the stage-specific checklist dimensions used for \textit{idea generation}, \textit{experiment design}, and \textit{paper synthesis}. Detailed definitions of the dimensions for each stage are provided in Figures~\ref{figure:dimension_idea}--\ref{figure:dimension_paper}.

\begin{figure}[htbp]
\centering
\begin{tcolorbox}[
  width=\linewidth,
  colback=black!7.5!white,
  colframe=black!80!white,
  title=Idea Generation Checklist Dimensions,
  fontupper=\scriptsize,
  fonttitle=\scriptsize,
  boxrule=0.4pt,
  sharp corners,
  before upper={\linespread{0.92}\selectfont},
]

\textbf{Problem.}
The concrete research problem, task, or setting that the research aims to address.

\par\smallskip
\textbf{Research Gap.}
The limitation, missing capability, or unresolved issue in existing work that leaves the research problem insufficiently addressed.

\par\smallskip
\textbf{Motivation.}
The observation, intuition, or hypothesis explaining why the proposed research direction or approach is expected to be useful or effective.

\par\smallskip
\textbf{Method.}
The core technical approach proposed to address the research problem, including its distinctive mechanisms and design choices.

\end{tcolorbox}

\caption{
Definitions of the checklist dimensions used for evaluating idea generation.
}
\label{figure:dimension_idea}
\end{figure}
\begin{figure}[htbp]
\centering
\begin{tcolorbox}[
  width=\linewidth,
  colback=black!7.5!white,
  colframe=black!80!white,
  title=Experiment Design Checklist Dimensions,
  fontupper=\scriptsize,
  fonttitle=\scriptsize,
  boxrule=0.4pt,
  sharp corners,
  before upper={\linespread{0.92}\selectfont},
]

\textbf{Experimental Objective.}
The research claims, hypotheses, or questions that the planned experiments are intended to test or establish.

\par\smallskip
\textbf{Experimental Setup.}
The datasets, benchmarks, models, systems, environments, and implementation choices used to conduct the experiments.

\par\smallskip
\textbf{Baselines and Comparisons.}
The baseline methods, comparison variants, reference systems, or conditions used to contextualize the proposed method.

\par\smallskip
\textbf{Evaluation Protocol.}
The metrics, evaluation splits, aggregation procedures, repeated runs, human evaluations, judges, verifiers, and uncertainty-reporting procedures used to assess experimental outcomes.

\par\smallskip
\textbf{Ablation and Analysis.}
The ablation studies, parameter or scale analyses, diagnostic breakdowns, and qualitative analyses used to examine the behavior and components of the proposed method.

\end{tcolorbox}

\caption{
Definitions of the checklist dimensions used for evaluating experiment design.
}
\label{fig:dimension_experiment}
\end{figure}
\begin{figure}[htbp]
\centering
\begin{tcolorbox}[
  width=\linewidth,
  colback=black!7.5!white,
  colframe=black!80!white,
  title=Paper Synthesis Checklist Dimensions,
  fontupper=\scriptsize,
  fonttitle=\scriptsize,
  boxrule=0.4pt,
  sharp corners,
  before upper={\linespread{0.92}\selectfont},
]

\textbf{Research Framing.}
How the paper introduces and frames the research problem, motivation, gap, and overall contribution.

\par\smallskip
\textbf{Content Selection and Emphasis.}
Which aspects of the research are selected for presentation and how emphasis is distributed across methods, experiments, findings, and other research content.

\par\smallskip
\textbf{Scientific Argumentation.}
How claims, evidence, and reasoning are connected to construct the scientific argument of the paper.

\par\smallskip
\textbf{Information Organization.}
How research content is structured and ordered across sections and how individual pieces of information are arranged into a coherent presentation.

\par\smallskip
\textbf{Interpretation and Positioning.}
How experimental findings are interpreted and how the resulting contributions are positioned relative to the research question and broader research context.

\end{tcolorbox}

\caption{
Definitions of the checklist dimensions used for evaluating paper synthesis.
}
\label{figure:dimension_paper}
\end{figure}

\subsection{Single-LLM-as-a-Judge Baseline}
\label{appendix:single}

This section provides additional details on the single-LLM-as-a-judge baseline used in our validation experiments. 
Unlike AutoHRS, which decomposes evaluation into checklist questions and aggregates their outcomes, the single-LLM judge directly compares the auto-researcher output with the corresponding human reference using the same stage-specific evaluation dimensions and assigns a single holistic reconstruction score. 
Figure~\ref{figure:single_llm_judge_prompt} presents the evaluation prompt for the idea generation stage as an example.
\begin{figure}[htbp]
\centering
\begin{tcolorbox}[
width=\linewidth,
colback=black!7.5!white,
colframe=black!80!white,
title=Prompt for Single-LLM-as-a-Judge Evaluation,
fontupper=\scriptsize,
fonttitle=\scriptsize,
boxrule=0.4pt,
sharp corners,
before upper={\linespread{0.92}\selectfont},
]

\texttt{[User Input]}
\par\smallskip
You are an expert ML/NLP research reviewer. Evaluate how faithfully a
candidate research idea reconstructs the corresponding human reference
idea.

Assess the candidate by comparing it directly with the reference idea
along the following dimensions:

\begin{enumerate}
\setlength{\itemsep}{0pt}
\setlength{\parskip}{0pt}

\item \textbf{Problem:}
Whether the candidate identifies the same research problem or task
setting as the reference.

\item \textbf{Research Gap:}
Whether the candidate identifies the same limitation or unresolved
issue in existing work.

\item \textbf{Motivation:}
Whether the candidate captures the same observation, intuition, or
rationale motivating the proposed research.

\item \textbf{Method:}
Whether the candidate proposes a method consistent with the core
technical approach and distinctive design decisions in the reference.

\end{enumerate}

Judge semantic similarity rather than exact wording. Do not reward the
candidate merely for belonging to the same research area. Differences in
the central research problem, gap, motivation, or proposed method should
reduce the evaluation score accordingly.

After considering all dimensions jointly, assign a single overall score
from 0 to 4 reflecting the reconstruction fidelity of the candidate idea:

\begin{itemize}
\setlength{\itemsep}{0pt}
\setlength{\parskip}{0pt}
\item \textbf{0}: No meaningful correspondence with the reference idea.
\item \textbf{1}: Limited correspondence; only a few broad elements match.
\item \textbf{2}: Partial reconstruction; some important elements match,
but substantial differences remain.
\item \textbf{3}: Strong reconstruction; most important research decisions
are consistent with the reference, with only minor differences.
\item \textbf{4}: Highly faithful reconstruction of the reference idea
across the major evaluation dimensions.
\end{itemize}

\textbf{Inputs:}\
Reference idea: {reference\_idea}\
Candidate idea: {candidate\_idea}

\par\smallskip
\textbf{Output format:}\
Return only a JSON object without markdown fences.

\par\smallskip
{\ttfamily
\noindent
\{\
\hspace*{1em}"score": <0, 1, 2, 3, or 4>,\
\hspace*{1em}"reason": "<brief justification of the overall score>"\
\}
\par
}

\end{tcolorbox}

\caption{
Prompt for the single-LLM-as-a-judge baseline, illustrated using the idea generation stage. The judge directly compares the candidate and human reference across the stage-specific evaluation dimensions and assigns a single holistic reconstruction score.
}
\label{figure:single_llm_judge_prompt}
\end{figure}

\subsection{Human Evaluation for Checklist Validation}
\label{appendix:human}

To validate the checklist-based evaluation against human judgments, we conduct a human evaluation on a randomly sampled subset of 20 examples from the evaluation corpus. For each example, annotators are provided with the output generated by the auto researcher together with the corresponding human reference output. Annotators are asked to directly assess the overall degree to which the auto-researcher output reconstructs the human reference, using a 5-point Likert scale from 0 to 4, where 0 indicates no meaningful reconstruction and 4 indicates highly faithful reconstruction.

Each example is independently evaluated by three human annotators. We assess inter-annotator agreement using the average-measures intraclass correlation coefficient, ICC(3,3), obtaining a value of 0.72, which indicates moderate agreement among annotators. We then average the three annotation scores for each example and use the resulting mean as the human judgment for correlation analysis with the checklist-based evaluation.

\section{Experiments}
\label{app:experiment}

\subsection{Auto-Researcher Configuration}
\label{appendix:auto_researcher_config}

\paragraph{Research Skills and Execution Environment}
We configure the research skills for the three evaluation stages by adapting the instructions from~\cite{YLL26}.
The skills specify the stage-specific inputs, task objectives, and expected output formats.
Each stage is executed independently using inputs extracted from the corresponding human reference paper.
We disable internet access, including web search and external retrieval, to prevent access to the target papers through online sources.
The target papers and human reference outputs are excluded from the agent-accessible workspace; only the designated stage inputs are provided.

\paragraph{Idea Generation and Selection}
We do not prescribe a fixed number of ideas per paper, allowing the auto researcher to determine how many candidate ideas to produce during each run.
Consequently, the number of generated ideas varies across papers.
We evaluate every generated idea using the reference-derived checklist and report results both across all generated ideas and for the highest-scoring idea per paper.
The best idea is selected post hoc according to its overall AutoHRS score, rather than through a selection rule specified in the research skill.
The checklist and its scores are used only for evaluation and are not provided to the auto researcher during generation.

\subsection{Robustness Across Agent Harnesses}
\label{appendix:harness}

\begin{table}[t!]
\centering
\caption{Agent harnesses and their underlying language models.}
\label{tab:harness}

\begin{tabular}{ll}
\toprule
\textbf{Harness} & \textbf{Underlying Model} \\
\midrule
Claude Code & Claude-Sonnet-4.5 \\
Gemini CLI  & Gemini-3.5-Flash \\
Codex       & GPT-5.2 \\
\bottomrule
\end{tabular}
\end{table}

\begin{figure*}[h]
\centering
\includegraphics[width=0.45\textwidth]{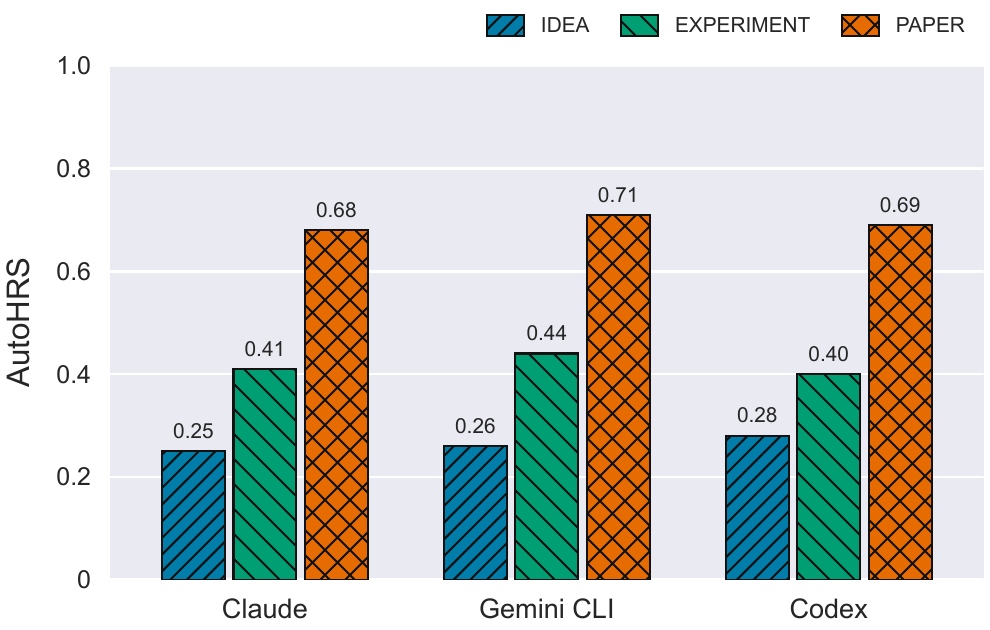}
\caption{\textbf{AutoHRS performance across different agent harnesses.}
Average AutoHRS scores of different harness-based auto researchers across the three research stages: \textit{idea generation}, \textit{experiment design}, and \textit{paper synthesis}.
}
\label{fig:harness}
\end{figure*}

To assess whether our findings hold across different agent harnesses, we replicate a subset of our experiments using Gemini CLI and Codex. 
To mitigate potential data contamination, we select underlying models whose documented knowledge cutoffs predate the earliest public release of the corresponding reference papers. Table~\ref{tab:harness} lists the harnesses and their underlying models.
As shown in Figure~\ref{fig:harness}, the three harnesses achieve comparable average AutoHRS scores, suggesting that our main findings are not specific to a particular harness. 
Given the substantial cost of running large-scale research trajectories, we therefore conduct the full evaluation using a single primary harness and use the remaining harnesses for robustness checks.

\subsection{Weight Selection}
\label{appendix:weight}

We initially compute the overall AutoHRS score for each stage by uniformly averaging its dimension-level scores. 
However, uniform weighting shows relatively limited correlation with human judgments, indicating that different dimensions contribute unequally to reconstruction quality. 
We therefore select stage-specific weights through a grid search over non-negative weight combinations that sum to one, choosing the combination that maximizes Spearman's correlation with human judgments. 
The resulting weights for each stage are reported in Table~\ref{table:weight}.

\begin{table}[!t]
\centering
\caption{\textbf{Stage-specific weights used in AutoHRS.}
The weights for each stage are selected to maximize Spearman's correlation with human judgments under the constraint that they are non-negative and sum to one.}
\label{table:weight}

\setlength{\tabcolsep}{3pt}
\scalebox{.95}{%
\begin{tabular}{p{0.3\linewidth} p{0.55\linewidth} c}
\toprule
\textbf{Stage} & \textbf{Dimension} & \textbf{Weight} \\
\midrule

\multirow{4}{*}{\textit{Idea Generation}}
& Problem & 0.20 \\
& Research Gap & 0.25 \\
& Motivation & 0.25 \\
& Method & 0.30 \\
\midrule

\multirow{5}{*}{\textit{Experiment Design}}
& Experimental Objective & 0.10 \\
& Experimental Setup & 0.20 \\
& Baselines and Comparisons & 0.30 \\
& Evaluation Protocol & 0.25 \\
& Ablation and Analysis & 0.15 \\
\midrule

\multirow{5}{*}{\textit{Paper Synthesis}}
& Research Framing & 0.10 \\
& Content Selection and Emphasis & 0.25 \\
& Scientific Argumentation & 0.30 \\
& Information Organization & 0.25 \\
& Interpretation and Positioning & 0.10 \\

\bottomrule
\end{tabular}%
}
\end{table}

\subsection{Granularity Examples}
\label{appendix:gra}

This section provides representative examples of research directions at different granularity levels used in our analysis.
As shown in Figure~\ref{figure:granularity_examples}, the research direction is progressively refined from a broad field to a fine-grained research scope while avoiding information that directly reveals the target problem, research gap, motivation, or method.

\begin{figure}[htbp]
\centering
\begin{tcolorbox}[
width=\linewidth,
colback=black!7.5!white,
colframe=black!80!white,
title=Example of Research Direction Granularity,
fontupper=\scriptsize,
fonttitle=\scriptsize,
boxrule=0.4pt,
sharp corners,
before upper={\linespread{0.92}\selectfont},
]

\textbf{Level 0: Broad Field}
\par\smallskip
Large Language Models

\par\medskip
\textbf{Level 1: Research Topic}
\par\smallskip
Reasoning in Large Language Models

\par\medskip
\textbf{Level 2: Focused Research Area}
\par\smallskip
Long-context reasoning in Large Language Models

\par\medskip
\textbf{Level 3: Fine-grained Research Scope}
\par\smallskip
Multi-hop reasoning over long textual contexts in Large Language Models

\end{tcolorbox}

\caption{
Example of the four research-direction granularity levels used in our analysis.
Each level progressively narrows the research scope while avoiding explicit information about the target problem, research gap, motivation, or method.
}
\label{figure:granularity_examples}
\end{figure}

\end{document}